\pdfoutput=1

\documentclass[11pt]{article}

\usepackage{acl}

\usepackage{times}
\usepackage{latexsym}
\usepackage{cancel}

\usepackage{etoolbox}
\robustify\citep
\robustify\citet

\usepackage[T1]{fontenc}

\usepackage[utf8]{inputenc}

\usepackage{microtype}

\usepackage[flushmargin,hang]{footmisc}

\usepackage{inconsolata}

\usepackage{graphicx}

\usepackage{microtype}
\usepackage{graphicx}

\usepackage{rycolab}
\usepackage{macros}

\renewcommand{\circ}[0]{{\cdot}}

\title{Syntactic Control of Language Models by Posterior Inference}

\author{
 Vicky Xefteri \qquad
 Tim Vieira \qquad
 Ryan Cotterell \qquad
 Afra Amini \\
 \texttt{\{\href{mailto:vxefteri@ethz.ch}{vxefteri}, \href{mailto:ryan.cotterell@ethz.ch}{ryan.cotterell}, \href{mailto:afra.amini@ethz.ch}{afra.amini}\}@ethz.ch} \quad\texttt{\href{mailto:tim.f.vieira@gmail.com}{tim.f.vieira@gmail.com}}\\
 \includegraphics[width=3cm]{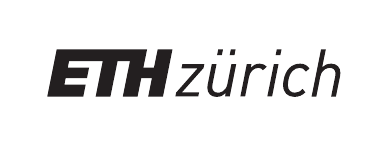}
 }

\begin{document}

\maketitle
\begin{abstract}

Controlling the syntactic structure of text generated by language models is valuable for applications requiring clarity, stylistic consistency, or interpretability, yet it remains a challenging task. In this paper, we argue that sampling algorithms based on the posterior inference can effectively enforce a target constituency structure during generation. Our approach combines sequential Monte Carlo, which estimates the posterior distribution by sampling from a proposal distribution, with a syntactic tagger that ensures that each generated token aligns with the desired syntactic structure. Our experiments with \gptt and \llama models show that with an appropriate proposal distribution, we can improve syntactic accuracy, increasing the F1 score from $12.31$ (\gptl) and $35.33$ (\llama) to about $93$ in both cases without compromising the language model's fluency. These results underscore both the complexity of syntactic control and the effectiveness of sampling algorithms, offering a promising approach for applications where precise control over syntax is essential.\looseness=-1

\faGithub\ {\tt\fontsize{9.65}{11}\selectfont\href{https://github.com/rycolab/syntactic-control}{github.com/rycolab/syntactic-control}}
\end{abstract}
\section{Introduction}

Syntactic control of generated text is crucial for many domain-specific applications of language models, where structural constraints, such as formality, grammatical correctness, or adherence to a given template, can significantly affect usability and readability. Despite recent advances, achieving fine-grained syntactic control remains a challenge for large-scale language models. More broadly, this challenge falls under the area of \emph{controlled} generation, which studies how to guide language models to produce text with desired properties. 

\begin{figure}[t] 
\noindent\textbf{User}: 
\begin{itemize}
\item \texttt{(S (NP (EX \?)) (VP (VBZ \?) (ADVP (RB
\?)) (NP (DT \?) (NN \?))))}
\end{itemize}

\noindent\textbf{System}:
\begin{itemize}
\item \texttt{(S (NP (EX {\color{PPink}There})) (VP (VBZ {\color{PPink}is}) (ADVP (RB
{\color{PPink}always})) (NP (DT {\color{PPink}a}) (NN {\color{PPink}chance}})))) 
{\color{gray}$[\potential=0.99, \plm=2.95e^{-13}]$}

\item \texttt{(S (NP (EX {\color{PPink}There})) (VP (VBZ {\color{PPink}is}) (ADVP (RB
{\color{PPink}never})) (NP (DT {\color{PPink}a}) (NN {\color{PPink}reason}})))) {\color{gray}$[\potential=0.99
,\plm=8.66e^{-14}]$}

$\vdots$
\end{itemize}

\noindent\textbf{\gptf baseline}:
\begin{itemize}
\item \texttt{(S (NP (EX {\color{PPink}There})) (VP (VBZ {\color{PPink}is}) (ADVP (RB {\color{PPink}clearly})) (NP (DT {\color{PPink}a}) {\color{POrange}\sout{(JJ {\color{PPink}big})}} (NN {\color{PPink}problem}))))} {\color{gray}$[\potential={\color{POrange}0},
\plm=1.90e^{-14}]$}
\end{itemize}

\caption{Example user--system interaction.  The user specifies a target syntactic structure as a Penn Treebank syntax tree with  {\tt\color{PPink}?} in the place of words. The system probabilistically fills in the missing words, such that the probability of the generated sentence is high under the LM ($\plm$) and the likelihood of the targeted syntactic structure is high under the syntactic analyzer ($\potential$). \gptf fails to generate a valid string, as it generates an additional adjective {\tt\color{POrange}{(JJ {\color{PPink}big})}}, making it impossible for the sequence of words to have the target syntax (i.e., $\potential=0$).\looseness=-1
}
\label{fig:example-interaction}
\vspace{-\baselineskip}
\end{figure}

Controlling the syntactic structure of sentences is crucial in some domains. For instance, maintaining stylistic consistency is important in fields such as legal, technical, and formal writing, as well as in creative writing, where sentence structure significantly contributes to tone and clarity. In educational settings, controlling syntactic complexity and ambiguity enables intelligent grammar tutoring in both native and foreign languages \citep{renduchintala-etal-2016-creating}. Syntactic control is also valuable for generating psycholinguistic stimuli, allowing studies on how syntax impacts cognitive load, memory, or reading time \citep{britton1982effects, syntaxcog}. More generally, it can support the generation of text that is easier to comprehend, for example, by avoiding deeply nested structures or long, ambiguous sentences \cite{kauchak-2013-improving,xu-etal-2015-problems}, and reduce semantic ambiguity (e.g., prepositional attachment ambiguities), which can leave readers confused or increase cognitive processing effort.
Finally, from an interpretability perspective, syntactic control provides a framework for analyzing how syntax affects the behavior of language models \citep{linzen-etal-2016-assessing, futrell-etal-2019-neural}.\looseness=-1

There are many techniques in the literature that perform controlled generation. One line of work in controlled generation\footnote{E.g., \citet{krause-etal-2021-gedi-generative, liu-etal-2021-dexperts, selfdebiasing, lew2023sequentialmontecarlosteering, zhao2024probabilisticinferencelanguagemodels}.} has focused on developing inference-time algorithms that control for qualities of generated text, such as topic, sentiment, and toxicity. 
Perhaps, the most relevant work focuses on generating code subject to the syntactic constraints of a \emph{programming language}.\footnote{E.g., \citet{scholak2021picard,poesia2022synchromesh,ugare2024syncode,loula2025iclr}.}  Our work, in contrast, considers controlled generation under the constraints of \emph{natural language} syntax.  Our setting is challenging because natural language does not have simple context-free rules and, thus, cannot be easily checked for violations during generation (e.g., to rule out poor choices from being sampled). Moreover, unlike programming languages, natural language syntax is much richer, inherently ambiguous, and harder to analyze, making it an interesting new challenge for controlled generation methods.\looseness=-1

An alternative to controlled generation is prompting instruction-tuned models. Recent studies \citep{sun-etal-2023-evaluating, ashok2024cg} have shown the promise of instruction-tuned LMs for following specific control targets, such as generating text on a particular topic or with a positive sentiment. A previously noted limitation of prompting instruction-tuned models is their difficulty with syntactic control, where prompting fails to guide models to produce text with the desired syntactic structure \citep{sun-etal-2023-evaluating, ashok2024cg}.

It is perhaps unsurprising that prompting alone is insufficient for reliably controlling syntax. 
Enforcing a global structure upon the generated string is inherently challenging. To improve syntactically controlled generation at inference time, we propose a sampling method that approximates the posterior distribution over strings generated by a language model under a target syntactic structure. Our approach is based on sequential Monte Carlo, an algorithm that estimates the posterior by drawing samples from a proposal distribution and weighting them by the likelihood that the given string follows a specific attribute, in our case, a syntax tree. In this paper, we use parsers-as-taggers\footnote{E.g., \citet{gomez-rodriguez-vilares-2018-constituent}, \citet{vacareanu-etal-2020-parsing}, and \citet{amini-cotterell-2022-parsing}.} to further guide the generation towards samples with higher likelihood.
We argue that taggers can provide effective guidance since they, by design, factorize the syntactic structure and align it with each word in the sequence.  Specifically, we use Tetratagger \citep{tetra}, which produces a constituency analysis through a clever linear encoding of the constituency tree that assigns a pair of tags to each word. While the original is trained on a masked language model, we also train Tetratagger with an autoregressive language model as its backbone to guide sampling.

Our experiments with \gptf \cite{openai2024gpt4technicalreport} align with previous findings \citep{sun-etal-2023-evaluating}, showing that the \gptf model has difficulty generating sentences with the desired constituency trees in zero- or few-shot settings. We then apply our method to \gptl \cite{radford_language_2019}, as well as to \llama models \cite{dubey2024llama3herdmodels}. Sentences generated using our method adhere much more closely to the target syntax than those produced without it. For instance, with a well-chosen proposal distribution, our method improves the F1 score of \gptl generations from $12.31$ to $93.69$ without degrading the model’s fluency. Moreover, our method can be applied to instruction-tuned models like \llama, achieving a similar F1 score with \gptl. These results demonstrate that controlled generation by posterior inference can make smaller models competitive with larger ones, like \gptf.

\section{Controlled Generation by Posterior Inference}

A \defn{language model} (\defn{LM}) $\plm$ is a probability distribution over strings $\str \in \alphabet^*$, where $\alphabet$ is the set of tokens in the vocabulary. Most state-of-the-art language models are factored into per-token conditional distributions. 
The probability of a string $\str$ under a language model is given by
\begin{align}\label{eq:lm}
\plm(\str) \defeq \plm(\eos \mid \str) \prod_{n=1}^{|\str|} \plm(\sym_n \mid \str_{<n}),
\end{align}
where $\str_{<n} \defeq \sym_1 \cdots \sym_{n-1}$, and $\eos \notin \alphabet$ is a distinguished, end-of-string token.

Language models contain a vast amount of prior knowledge about what constitutes well-formed and fluent natural language. However, they have limited inherent understanding of external constraints or specific control targets. In the context of this work, the language model serves as a \defn{prior} probability $\plm(\str)$ that a string \str is desirable.\footnote{One may freely modify this prior, e.g., applying top-$p$ filtering, annealing, or including instructions, to account for preferences that are independent of the target syntax.} In order to \emph{control} generation, we define an additional factor, the \defn{likelihood} $\potential(\bt \mid \str)$ that \str possesses the desired attribute $\bt$.\footnote{We use $\potential$ to denote the likelihood to prevent confusion with the conditional distribution over tokens used in \cref{eq:lm}.}  Together, the prior and the likelihood define a \defn{posterior} distribution $\posterior(\str \mid \bt)$ that accounts for the inherent uncertainties in each distribution:\looseness=-1
\begin{subequations}
\begin{align} \label{eq:bayes}
\annotateOver{\posterior(\str \mid \bt)}{posterior}
&= \frac{\annotateOver{\plm(\str)}{prior}\  \annotateOver{\potential(\bt \mid \str)}{likelihood}}{\annotateUnder{\Z(\bt)}{evidence}}
\\
\Z(\bt) &\defeq \sum_{\str \in \alphabet^*} \plm(\str)\, \potential(\bt \mid \str)
\end{align}
\end{subequations}
Note that $\Z(\bt)$ is the probability that any string generated by $\plm$ has the attribute $\bt$. 
Unfortunately, the exact computation of $\Z(\bt)$ and exact sampling from $\posterior(\str \mid \bt)$ are generally intractable. 
However, we will present principled, practical approximations in \Cref{sec:is,sec:smc}.

\paragraph{Likelihood.}
Note that $\potential(\bt \mid \str)$ is defined by a neural model that has been trained to approximate the true association between the target attribute $\bt$ and strings \str. 
In the case of categorical labels,\footnote{E.g., sentiment, topic, and toxicity \citep[][\emph{inter alia}]{fudge, liu-etal-2021-dexperts, amini2023structured}.} $\potential(\bt \mid \str)$ may be a probabilistic classifier trained to predict a label $\bt$ for each string \str (e.g., $\bt$ could be a positive or negative sentiment judgment, and $\potential(\bt \mid \str)$ can be learned from a sentiment classification corpus).  However, in this paper, $\bt$ is a user-specified syntax tree where the internal nodes of a syntax tree are provided, but the words are left for the LM to provide (see \cref{fig:example-interaction} for an illustration). Thus, $\potential(\bt \mid \str)$ is a rich probabilistic model that assigns conditional probabilities to syntactic analyses $\bt$ for each string $\str$;\footnote{Note that our approach differs from other syntax-constrained generation methods \citep[e.g.,][] {shin2021constrained, scholak2021picard, poesia2022synchromesh, ugare2024syncode} because our control signal $\potential(\bt \mid \str)$ (a) is probabilistic, (b) is a complex model rather than a simplistic context-free grammar.} see \cref{fig:example-interaction} for an illustration.
We describe our syntactic likelihood model in \cref{sec:tetra}. In general, the likelihood doesn't necessarily need to be a conditional probability distribution. We can view our abstract problem as 
\begin{subequations}
\begin{align} \label{eq:bayes-potential}
\annotateOver{\posterior(\str)}{posterior}
&= \frac{\annotateOver{\plm(\str)}{prior}\  \annotateOver{\potential(\str)}{potential}}{\annotateUnder{\Z}{evidence}}
\\
\Z &\defeq \sum_{\str \in \alphabet^*} \plm(\str)\, \potential(\str),
\end{align}
\end{subequations}

where $\potential(\str) \ge 0$ (for all $\str$) is often called a \defn{potential} function, rather than a likelihood function.

\subsection{Importance Sampling} \label{sec:is}
\defn{Importance sampling} (\defn{IS}) offers a practical approximation to the posterior distribution $\posterior(\str)$.\footnote{We refer the reader to \citet{chatterjee2017samplesizerequiredimportance} for a thorough analysis of importance sampling.} Given $\potential$ and a sample size $M$, importance sampling works as follows: 
\begin{enumerate}[noitemsep,topsep=0pt,parsep=0pt,partopsep=0pt,leftmargin=*]
\item Sample $\str^{(1)}, \ldots, \str^{(M)} \overset{\text{i.i.d.}}{\sim} \proposal$ where $\proposal$ is a \defn{proposal distribution}, which is another language model that should approximate the posterior.\footnote{%
Ideally, we want a proposal distribution $\proposal \approx \posterior$. However, for correctness, it is sufficient that whenever $\proposal(\str) = 0$, we also have $\plm(\str)\,\potential(\str) = 0$.\label{proposal-AC}
If this condition is not satisfied, then any string $\str$ with $\plm(\str)\,\potential(\str)\neq 0$ but $\proposal(\str) = 0$, will never be sampled, leading to a false zero in the posterior approximation that cannot be resolved by taking more samples.\looseness=-1
} We refer to each of these samples as a \defn{particle}.

\item Evaluate the potential $\potential(\str^{(m)})$, and compute the particle \defn{weight} $\uweight(\str) \defeq \frac{\plm(\str)}{\proposal(\str)} \potential(\str)$.  Let $\uweight^{(m)} = \uweight(\str^{(m)})$.

\item Compute the posterior approximation:
\begin{subequations}
\begin{align}\label{eq:appr_distrb}
\!\!\!\!\pest(\str)
&\defeq \frac{1}{M \, \zest}\sum_{m=1}^M \uweight^{(m)} \mathbbm{1} \{\str = \str^{(m)} \}
\\
\zest 
&\defeq \frac{1}{M} \sum_{m=1}^M \uweight^{(m)}
\end{align}
\end{subequations}
\end{enumerate}

\noindent Now, to draw (approximate) samples, we draw strings \str from the posterior approximation $\pest$, which is efficient as there are at most $M$ strings with nonzero probability.
Importance sampling comes with the following two guarantees:
\begin{itemize}
\item $\zest$ is an unbiased estimate of $\Z$ 
\item $\pest(\str)$ is consistent estimate of $\posterior(\str)$, i.e., it converges as $M \to \infty$.
\end{itemize}

\paragraph{\emph{Sequential} importance sampling (SIS).}
Sequential importance sampling \citep[e.g.,][]{doucet2001sequential} is an implementation of importance sampling tailored for sequences, where samples are drawn from the conditional proposal distribution and weights are computed incrementally at each step. The specific details are given in \cref{alg:sis}, but it is equivalent to importance sampling with the addition of the boolean $\activeParticle$ that tracks which particles are still incomplete (not yet $\eos$-terminated).

\begin{algorithm}[t] 
\caption{Sequential importance sampling}
\begin{algorithmic}[1]
\small
\Procedure{SIS}{$\plm, \proposal, \potential, M$}

\For{$m = 1 \ldots M$}    \Comment{$M$ number of particles}
  \State $(\str^{(m)}, \uweight^{(m)}, \activeParticle^{(m)}) \gets (\varepsilon, 1, \texttt{true})$
\EndFor

\While{$\exists m \!\in\! 1 \ldots M \colon \activeParticle^{(m)}$} 

\For{$m=1 \dots M\colon \neg \activeParticle^{(m)}$} \Comment{Incomplete particles}

\State $\sym' \sim \proposal(\cdot \mid \str^{(m)})$
\If{$\sym' = \eos$}   \Comment{Complete the particle}
  \State $\activeParticle^{(m)} \gets \texttt{false}$ 
  \State $\uweight^{(m)} \gets \uweight^{(m)} \!\cdot\! \frac{\plm(\sym' \mid \str^{(m)})}{\proposal(\sym' \mid \str^{(m)})} \potential(\str^{(m)})$ \label{line:sis-weight-complete}
\Else
  \State $\uweight^{(m)} \gets \uweight^{(m)} \!\cdot\! \frac{\plm(\sym' \mid \str^{(m)})}{\proposal(\sym' \mid \str^{(m)})}$ \label{line:sis-weight-incomplete}  
  \State $\str^{(m)} \gets \str^{(m)} \circ \sym'$
\EndIf
\EndFor

\EndWhile

\State $\zest \gets \frac{1}{M} \sum_{m=1}^M \uweight^{(m)}$

\State $\pest(\str) \gets \totalWeight^{-1} \sum_{m=1}^M \uweight^{(m)} \mathbbm{1}\{ \str \!=\! \str^{(m)}\}$

\State \Return $(\zest, \pest)$ 

\EndProcedure

\end{algorithmic}
\label{alg:sis}
\end{algorithm}

\subsection{Sequential Monte Carlo}
\label{sec:smc}

The drawback of importance sampling is that it often allocates computation (i.e., sampling budget) poorly because it may sample many particles that show early signs of being poor samples.  Thus, rather than sampling complete strings from the proposal $\proposal$, we seek to \emph{incrementally} evaluate and prioritize samples as they evolve.\footnote{Therefore, we assume that the proposal $\proposal$ factors the same way that the prior $\plm$ does \cref{eq:lm}, i.e., it provides efficient methods to evaluate conditional probabilities $\proposal(\sym' \mid \str)$.}
\defn{Sequential Monte Carlo} \citep[\defn{SMC};][]{doucet2001sequential} augments importance sampling with two key ingredients: \emph{shaping} and \emph{resampling}.

\paragraph{Shaping.}  The idea behind shaping is that while we are generating a string, we should assess the quality of the partially completed particle with respect to the target posterior distribution using a shaping function. A \defn{shaping function}\footnote{Note that \citet{zhao2024probabilisticinferencelanguagemodels} call them \emph{twisting} functions.} $\shape\colon \alphabet^* \to \mathbb{R}_{\ge 0}$ approximates the expected future potential $\shape^*(\str)$ of the partially completed string $\str$:
\begin{align}
\shape(\str) \approx \shape^*(\str) \defeq \E[\yvar \sim \plm]{\potential(\yvar) \,\middle|\, \yvar \succeq \str }
\label{eq:optimal-shape}
\end{align}
where $\yvar \succeq \str$ denotes the event that the string-valued random variable $\yvar$, distributed according to the language model $\plm$, has $\str$ as a prefix.
Note that the \emph{exact} conditional probability of $\sym'$, a token or $\eos$, given the prefix $\str$ under the posterior distribution is in fact
\begin{align}
\shape^*(\sym' \mid \str) = \frac{\shape^*(\str \circ \sym')}{\shape^*(\str)},
\end{align}
and $\shape^*(\varepsilon) = \Z$. The optimal proposal distribution $\proposal^*$ is $\proposal^*(\sym' \mid \str) = \shape^*(\sym' \mid \str)$, as it draws exact samples according to auto-regressive factorization of the posterior distribution, i.e., $\posterior(\sym' \mid \str)$.\footnote{Note that the shaping function is akin to a \emph{heuristic} function in search \citep{pearl1984heuristics}; however, the specifics of what makes a heuristic function admissible (i.e., correct to use) differ from those that make a shaping function admissible.  We also note that the exact shaping function $\shape^*$ is closely related to the backward probabilities of hidden Markov models.
} Unfortunately, computing $\shape^*$ is, in general, intractable; however, SMC methods allow for approximating $\shape^*$ with a shaping function $\shape$, provided they are \defn{admissible} with respect to the target $\potential$, i.e., $\forall \str, \str' \in \alphabet^*\colon \plm(\str) \, \shape(\str) = 0 \Longrightarrow \potential(\str \circ \str') = 0$.
This condition ensures the particles cannot be incorrectly killed off (i.e., assigned weight zero) by the shaping function. We discuss our proposed approximation in \Cref{sec:tagen}.

\paragraph{Algorithm.}
SMC reallocates computational resources to the more promising particles based on their shaped weights such that many of the nice statistical properties of the importance sampling method are preserved.  Below, we provide a brief conceptual overview of the SMC algorithm, but refer to the pseudocode in \cref{alg:smc} for a complete technical description. The algorithm begins by initializing $M$ identical particles. Each $1 \le m \le M$ particle is represented by a string $\str^{(m)}$, a weight $\uweight^{(m)}$, and a boolean $\activeParticle^{(m)}$ that indicates if it has been completed (i.e., reached $\eos$).   The algorithm runs for $N$ steps, where $N$ is the maximum number of tokens we are willing to consider under the posterior distribution.\footnote{In our setting, we know the specific number of words that the string must have, so we use that value for $N$.}  For each particle $m$, while $\activeParticle^{(m)}$ is true, its weight $\uweight^{(m)}$ is a shaped estimate; however, once the particle is complete $\activeParticle^{(m)}$ is false, the weight will equal that of importance sampling (i.e., $\potential(\str^{(m)})$), as the product of the shaped estimates from the first step to this final step cancel each other out.\footnote{Consider $\str^{(m)} = \sym_1 \sym_2 \cdots \sym_{N-1} \sym_N$, its final weight is \\
$\shape(\varepsilon)
\frac{\shape(\sym_1)}{\shape(\varepsilon)}
\frac{\shape(\sym_1 \sym_2)}{\shape(\sym_1)}
{\cdots}
\frac{\shape(\sym_1 \cdots \sym_N)}{\shape(\sym_1 \cdots \sym_{N-1})}
\frac{\potential(\sym_1 \cdots \sym_N)}{\shape(\sym_1 \cdots \sym_N)}
\!=\! \potential(\str)$.
}

At each step, each particle $\str^{(m)}$ is extended by an additional token
by sampling from $\sym' \sim \proposal(\cdot \mid \str^{(m)})$.  If $\sym'$ is $\eos$, it is treated specially by updating its $\activeParticle^{(m)}$ and finalizing its weight $\uweight^{(m)}$; otherwise, we update $\str^{(m)}$ by appending the new token $\sym'$ to it and we update its weight $\uweight^{(m)}$ by multiply the shaping ratio, i.e., $\frac{\shape(\str^{(m)} \circ \sym')}{\shape(\str^{(m)})}$.

Once all particles have advanced to the next step, we may \emph{resample} (bootstrap) the particles, i.e., sample $M$ samples with replacement proportionally to their weights ($\uweight^{(m)}$). The resampled set of $M$ particles replaces the existing particles. We set the weights of the resampled particles equal to the average weight, as this choice preserves unbiasedness of $\zest$.\footnote{This resampling strategy is known as \emph{multinomial resampling} in the SMC literature \citep{doucet2001sequential}.} Most importantly, the lower-weight particles are less likely to be selected relative to the higher-weight particles.   This means the more promising particles, i.e., those with higher weights, are more likely to be replicated, and the later steps will extend their replicas.

Notice, however, that, unlike importance sampling, the samples produced by this method are no longer independent due to their shared history. This has the downside of producing a low-diversity sample, but it has the upside that the samples produced tend to be more representative of the posterior distribution.  To mitigate the issue of overly dependent samples, the resampling step is typically not performed at every step, but only when necessary.  A common criterion for deciding when to resample is the \defn{effective sample size} $\ess$ (defined on \cref{line:ess-def})\footnote{$\ess$ translates a set of weighted samples into an equivalent number of unweighted samples in relation to the variance reduction. For ordinary, unweighted Monte Carlo, $M$ samples reduce the variance of the estimator by roughly a factor of $1/M$. In contrast, for importance sampling, the variance reduction is roughly reduced by $1/\ess$ \citep{Martino_2017}.}
Resampling is triggered only when $\ess$ falls below a predefined threshold $\tau M$.

\begin{algorithm}[t] 
\caption{Sequential Monte Carlo}
\begin{algorithmic}[1]
\small
\Procedure{SMC}{$\plm, \proposal, \potential, \shape, M, \tau$}

\For{$m = 1 \ldots M$}    \Comment{$M$ number of samples}
  \State $(\str^{(m)}, \uweight^{(m)}, \activeParticle^{(m)}) \gets (\varepsilon, \shape(\varepsilon), \texttt{true})$
\EndFor

\While{$\exists m \in 1 \ldots M\colon \activeParticle^{(m)}$}
\For{$m=1 \dots M\colon \neg \activeParticle^{(m)}$} \Comment{Incomplete particles}

\State $\sym' \sim \proposal(\cdot \mid \str^{(m)})$
\If{$\sym' = \eos$}   \Comment{Complete the particle}
  \State $\activeParticle^{(m)} \gets \texttt{false}$ 
  \State $\uweight^{(m)} \gets \uweight^{(m)} \frac{\plm(\sym' \mid \str^{(m)})\potential(\str^{(m)})}{\proposal(\sym' \mid \str^{(m)})\shape(\str^{(m)})}$ \label{line:smc_weight-final}  \Comment{Final}
\Else
  \State $\uweight^{(m)} \gets \uweight^{(m)} \frac{\plm(\sym' \mid \str^{(m)})\shape(\str^{(m)} \circ \sym')}{\proposal(\sym' \mid \str^{(m)})\shape(\str^{(m)})}$ \Comment{Shaped} \label{line:smc_weight-shaped}  
  \State $\str^{(m)} \gets \str^{(m)} \circ \sym'$
\EndIf
\EndFor

\State $(\str^{(\cdot)}, \uweight^{(\cdot)}, \activeParticle^{(\cdot)})
\gets \textsc{Resample}(\str^{(\cdot)}, \uweight^{(\cdot)}, \activeParticle^{(\cdot)}, 
\tau
)$\!\!\!\!

\EndWhile

\State $\zest \gets \totalWeight/M$ 
\State $\pest(\str) \gets \totalWeight^{-1} \sum_{m=1}^M \uweight^{(m)} \mathbbm{1}\{ \str \!=\! \str^{(m)}\}$

\State \Return $(\zest, \pest)$ 

\EndProcedure

\vspace{.5\baselineskip}
\Procedure{Resample}{$\str^{(\cdot)}, \uweight^{(\cdot)}, \activeParticle^{(\cdot)}, 
\tau$}

\State $\totalWeight \gets \sum_{m=1}^M \uweight^{(m)}$

\State $\ess \gets \totalWeight^2 / \left(\sum_{m=1}^M \big(\uweight^{(m)}\big)^2 \right)$\label{line:ess-def}

\If{$\ess < \tau \!\cdot\! M$}  \Comment{Resample}

\State $\overline{\str}^{(\cdot)} \gets \str^{(\cdot)};\ \tmpWeight^{(\cdot)} \gets \uweight^{(\cdot)}$ \Comment{Temporary copy}

\For{$m = 1 \ldots M$}
\State $R \sim \mathrm{Categorical}(\totalWeight^{-1}\langle \tmpWeight^{(1)}, \ldots, \tmpWeight^{(M)}\rangle)$
\State $(\str^{(m)}, \uweight^{(m)}, \activeParticle^{(m)}) \gets (\overline{\str}^{(R)}, \totalWeight/M, \activeParticle^{(R)})$
\EndFor
\EndIf

\State \Return $(\str^{(\cdot)}, \uweight^{(\cdot)}, \activeParticle^{(\cdot)})$

\EndProcedure

\end{algorithmic}
\label{alg:smc}
\end{algorithm} 

\paragraph{Guarantees.}
SMC has the same guarantees as importance sampling.  We also note that shaping is \emph{ignored} if resampling is disabled by setting $\tau=0$, making the SMC algorithm's samples equivalent to importance sampling, and the procedure equivalent to sequential importance sampling.

\section{Tetratagger Likelihood}
\label{sec:tetra} 
In this section, we describe the syntactic likelihood model $\potential(\bt \mid \str)$, based on tetratagger.
A tetratagger is a neural constituency parser that works by assigning two tags to each word in the sentence, except for the last word, which is assigned one tag. The tags assigned to a word represent how this word and its parent (which is an intermediate node) are situated in the constituency tree. We denote the set of tags by $\tags$, which consists of \emph{four} tag types,\footnote{Two of the tag types encode information about the leaf nodes in the constituency tree and the other two encode information about the internal nodes. The four tags can be enhanced with labels for labeled trees; See \citep[\S 3.5;][]{tetra}.} hence the name \emph{tetra}tagger. Given a string of $L$ words $\str = \word_1 \word_2 \cdots \word_L$,\footnote{Here, we slightly abuse the notation $\str$ to represent both a sequence of tokens and a sequence of words, allowing flexibility in how we segment the sequence.} its corresponding binarized constituency contains $2L\!-\!1$ nodes.\footnote{Binarization transforms a constituency tree to a binary tree, where each node has at most two children. We refer the reader to \citep[\S 3.5;][]{tetra} for an explanation of how Tetratagger handles non-binary trees.} Tetratagger encodes the syntactic structure by assigning two tags to each word in the sequence. Therefore, $\potential$ models two conditional probability distributions over tags $\tags$ per word, except the last word to which we only assign one tag. For notational convenience, we assign an extra dummy tag to the last word with probability $1$, such that now each word is assigned exactly two tags. Given $\str$, the probability of a tag sequence $\bt$ under $\potential$ is

\begin{align}\label{eq:tetra}
\hspace{-5pt}\potential(\bt \!\mid\! \str) \defeq 
\smashoperator{\prod_{\ell=1}^{L}} \potential(t_{2\ell-1} \!\mid\! \str) 
\potential(t_{2\ell} \!\mid\! \str)
\end{align}

\paragraph{Modeling.}
The Tetratagger's conditional distributions $\potential(t_{2\ell-1} \mid \str)$ and $\potential(t_{2\ell} \mid \str)$ are defined by a pair of linear transformations on top of the existing transformer-based LM. More specifically, we first pass $\str$ to a language model and extract the $d$-dimensional latent representation $\lm(\word_\ell \mid \str) \in \mathbb{R}^d$ of $\word_\ell$ from the last layer of the language model. Next, we apply each transformation $\projOdd, \projEven \in \mathbb{R}^{\tags \times d}$ to this representation. Finally, we apply softmax to predict the distribution over the tags:%
\begin{subequations}
\begin{align} \label{eq:tetramodel}
\!\!\potential(t_{2\ell-1} \mid \str) &\defeq \mathrm{softmax}\big(\projOdd \, \lm(\word_\ell \mid \str) \big)_{t_{2\ell-1}}\\
\potential(t_{2\ell} \mid \str) &\defeq \mathrm{softmax}\big(\projEven \, \lm(\word_\ell \mid \str) \big)_{t_{2\ell}}
\end{align}
\end{subequations}
\noindent Given a dataset of strings and their corresponding ground-truth tetratags, i.e., a dataset of $(\str, \bt)$ pairs, we learn $\projOdd$ and $\projEven$ by maximizing the conditional log-likelihood of the dataset.\looseness=-1

\paragraph{Tokens vs.\@ words.} While the language model in \Cref{eq:tetramodel} is assumed to represent \emph{words} in a string, language models operate over \emph{tokens} and not words. Therefore, following \citet{tetra}, we set $\lm(\word_\ell \mid \str)$ to be the representation of the \emph{last token} of $\word_\ell$ extracted from the last layer of the language model.\looseness=-1

\section{Autoregressive Tetratagger Shaping}
\label{sec:tagen}
We now introduce an \emph{autoregressive} Tetratagger, which we use as a shaping function ($\shape$).  We first transform the desired constituency tree to a sequence of tags $\bt = t_1t_2\cdots t_{2L-1}$, where $L$ is the number of words in the tree's yield $\str$. For simplicity, we will assume for now that each word consists of a single token (i.e., $L \!=\! N$); We will explain at the end of this section how to support multi-token words. We assume that the Tetratagger is using an autoregressive LM backbone by factorizing \Cref{eq:tetra}. Thus, we define the autoregressive tetratagger probability as\looseness=-1

\begin{align}\label{eq:tetraauto}
\hspace{-5pt}\shape(\bt \mid \str) \defeq 
\smashoperator{\prod_{n=1}^{N}} \shape(t_{2n-1} \!\mid\! \str_{\leq n}) \shape(t_{2n} \!\mid\! \str_{\leq n}),
\end{align}
The crucial difference from the likelihood $\potential$ (\cref{eq:tetra}) is that when we predict the two tags at position $n$, we only have access to the string up to that point ($\str_{\leq n}$), rather than the complete string ($\str$).  This makes $\shape$ an effective shaping function as it can be efficiently evaluated as the words of the sentence are being generated left to right.\footnote{The ratio $\frac{\shape(\str^{(m)} \circ \sym')}{\shape(\str^{(m)})}$ on \cref{line:smc_weight-shaped} at the $n$-th step then simplifies to $\shape(t_{2n-1} \!\mid\! \str^{(m)} \circ \sym') \shape(t_{2n} \!\mid\! \str^{(m)} \circ \sym')$.}
While one might argue that bi-directional information may be critical for accurately predicting syntactic structure, we found that our autoregressive tetratagger is a decent parser (see \Cref{sec:trainingshaping}). Note that at the final step where we generate the $\eos$ token (\Cref{line:smc_weight-final}), we calculate the weights by using the likelihood $\potential(\bt \mid \str)$, as it is now a complete sentence.\footnote{Note that if we sample a different token at the $(N+1)\textsuperscript{th}$ step, we instead force the $\eos$ token to be generated since our generated sentence has a length-constraint.}  Lastly, we mention that $\shape$ is parameterized in the same way as $\potential$ (see \cref{eq:tetramodel}); thus, it can be trained in precisely the same way.  We provide additional training details in \cref{sec:trainingshaping}.

\paragraph{Multi-token words.} We again highlight the fact that Tetratagger assigns tag probabilities to the last token of each word. Therefore, in cases where the sampled token from the proposal LM is not the last token of the word, we keep sampling from the LM until we hit the last token. To detect word boundaries, we use a heuristic algorithm, where we decode the next token and observe whether the next token starts a new word.

\begin{table}[t] 
    \centering
    \begin{tabular}{@{}lrrr@{}}
        \toprule
         \textbf{Metrics} & \textbf{0-shot} & \textbf{5-shot} & \textbf{1-shot (gold)}  \\
        \midrule
        correct length & $42.05$ & $43.37$ & $65.89$ \\
        exact match & $17.55$ & $19.54$ & $31.78$ \\
        structure match & $20.53$ & $25.16$ & $44.37$  \\
        F1  & $47.23$ & $53.27$ & $70.17$ \\
        $\log \potential$ & $-\infty$ & $-\infty$ & $-17.41$ \\
        \bottomrule
    \end{tabular}
    \caption{Results of \gptf with $0$-shots, $5$-shots and $1$-shot of gold example on our evaluation dataset. We observe \gptf does fails to generate sentences with the correct syntactic structure in most cases, and achieves F1 score of only $53.27$ with $5$ examples. Even when the model has access to the ground-truth sentence ($1$-shot (gold)) it only achieves F1 score of $70.17$ on the dataset.\looseness=-1}
    \label{tab:gpt4results}
\end{table} 
\section{Experiments} \label{sec:dataset}
\paragraph{Dataset.} We compile a dataset of constituency trees from a subset of human-generated sentences originally produced by \citet{chen-etal-2019-controllable}. The dataset is built using paraphrase pairs from the ParaNMT dataset \cite{wieting-gimpel-2018-paranmt}. Each data point consists of a semantic input, a syntactic input, and a reference. The reference shares the same semantic content as the semantic input and mirrors the syntactic structure of the syntactic input. For our experiments, we use the references, which are human-generated sentences that exhibit syntactic variation. We then use the Berkeley Neural Parser \cite{kitaev-etal-2019-multilingual, kitaev-klein-2018-constituency} to parse these sentences to constituency trees. The leaf nodes of the trees that correspond to the words of the initial sentences are replaced with question marks ("?"), as depicted in \Cref{fig:example-interaction}. Our dataset consists of a diverse set of trees as shown in \Cref{tab:data_analysis}.

\paragraph{Prompt formulation.} For instruction-tuned models, we first experimented with various prompt formulations and ultimately selected the best-performing one shown in \Cref{sec:appendix_prompt}. To include the constituency trees in this prompt for few-shot settings, we need to linearize the tree. This is done using the bracketing representation of the trees and by replacing the leaf nodes with question marks as mentioned above. Note that for non-instruction models, the prompt is empty.

\paragraph{Evaluation.}
For evaluation of the generated sentences, we parse them to constituency trees and Tetratags. We compute the following metrics:
\begin{itemize}
    \item \textbf{correct length} is the percentage of generated sentences that match the exact word count specified by the constituent tree.
    \item \textbf{exact match} is the percentage of sentences whose generated parse trees match the desired parse trees exactly, both in structure and labels. 
    \item \textbf{structure match} is the percentage of sentences whose generated parse trees match the desired parse trees in structure (ignoring labels).
    \item \textbf{F1} is the mean bracketing F1 score comparing the desired constituency trees with the constituency trees of the generated sentences.
    \item $\log\potential$ is the median\footnote{We compute the $\log$ of the median of the exponentiated $\log\potential$ values, rather than using the mean, to prevent $-\infty$ values from skewing the results.} log-likelihood of the original Tetratagger, as defined by \Cref{eq:tetra} and trained with BERT by \citet{tetra} across all generated sentences of dataset.
    \item $\log \plm$ is the average log-prior probability of the generated string by the language model used.
    \item \textbf{diversity ($n$-gram)} is the average number of distinct $n$-grams in the sentences generated from a given tree, normalized by the total length of all sentences.
\end{itemize}

\subsection{\gptf Performance} \label{sec:tc} 
We first examine how challenging it is for state-of-the-art instruction-tuned language models to follow specific syntactic structures. Specifically, we assess how effectively \gptf can generate sentences that match the target syntactic structure, both in zero-shot and few-shot settings. The results are reported in \Cref{tab:gpt4results}. First, we observe that only $42.05\%$ and $43.37\%$ of \gptf generations have the desired length using $0$ and $5$ shots, respectively. By manually inspecting the generations, we noticed that \gptf tends to add adjectives, adverbs, or commas to sentences even though the corresponding syntax does not include one. For example, for the syntax \texttt{(S (NP (EX ?)) (VP (VBZ ?) (ADVP (RB ?)) (NP (DT ?) (NN ?))))}, \gptf produces the sentence ``{\tt There is clearly a {\color{POrange}big} problem}''. This leads to sentences with incorrect lengths. \looseness=-1

In all experimental setups, as reflected in \Cref{tab:gpt4results}, \gptf struggles to generate sentences with the desired structure. \gptf achieves F1 score of $47.23$ and $53.27$ with $0$-shot and $5$-shot respectively. To explore how far we can improve upon $0$-shot and $5$-shot performance, we experiment with including an example of a sentence with the exact desired constituency tree in the prompt and evaluate the model's performance in a $1$-shot (gold) setting. Surprisingly, even with the gold sentences included in the prompt, the model fails to consistently copy the sentence from the prompt into the output, achieving only an F1 score of $70.17$. Overall, the results indicate that generating sentences with a specific syntactic structure remains a challenging task, even for state-of-the-art language models.\looseness=-1

\subsection{Training Tetratagger Shaping Functions} \label{sec:trainingshaping}

We train Tetratagger using the \gptt and \llama models as the language model backbone. We later use these taggers to control the decoding process. Importantly, we match the LM backbone of the tagger with the language model used for generating text. This is essential to align the tokenization scheme of Tetratagger with the language model we use to generate text. Both Tetratagger models were trained for two epochs on the Penn Treebank dataset \cite{marcus-etal-1993-building}, using the cross-entropy loss.

We report the accuracy of our autoregressive Tetratagger in predicting correct tags for leaf nodes (Leaf Acc.) and internal nodes (Internal Acc.). Furthermore, we convert the sequence of predicted tags to constituency trees and compute the bracketing F1 score. Our autoregressive Tetrataggers with \gptl and \llama achieve an F1 score of $68.75$ and $71.79$, Leaf Acc. of $93.52$ and $94.45$, and Internal Acc. of $81.53$ and $82.96$ respectively, showing that we can achieve a reasonable performance even though we do not include any bidirectional information and thus our tagger does not have access to the whole sequence when predicting each tag.\looseness=-1

\begin{figure}[t]
\centering
\includegraphics[width=\columnwidth]{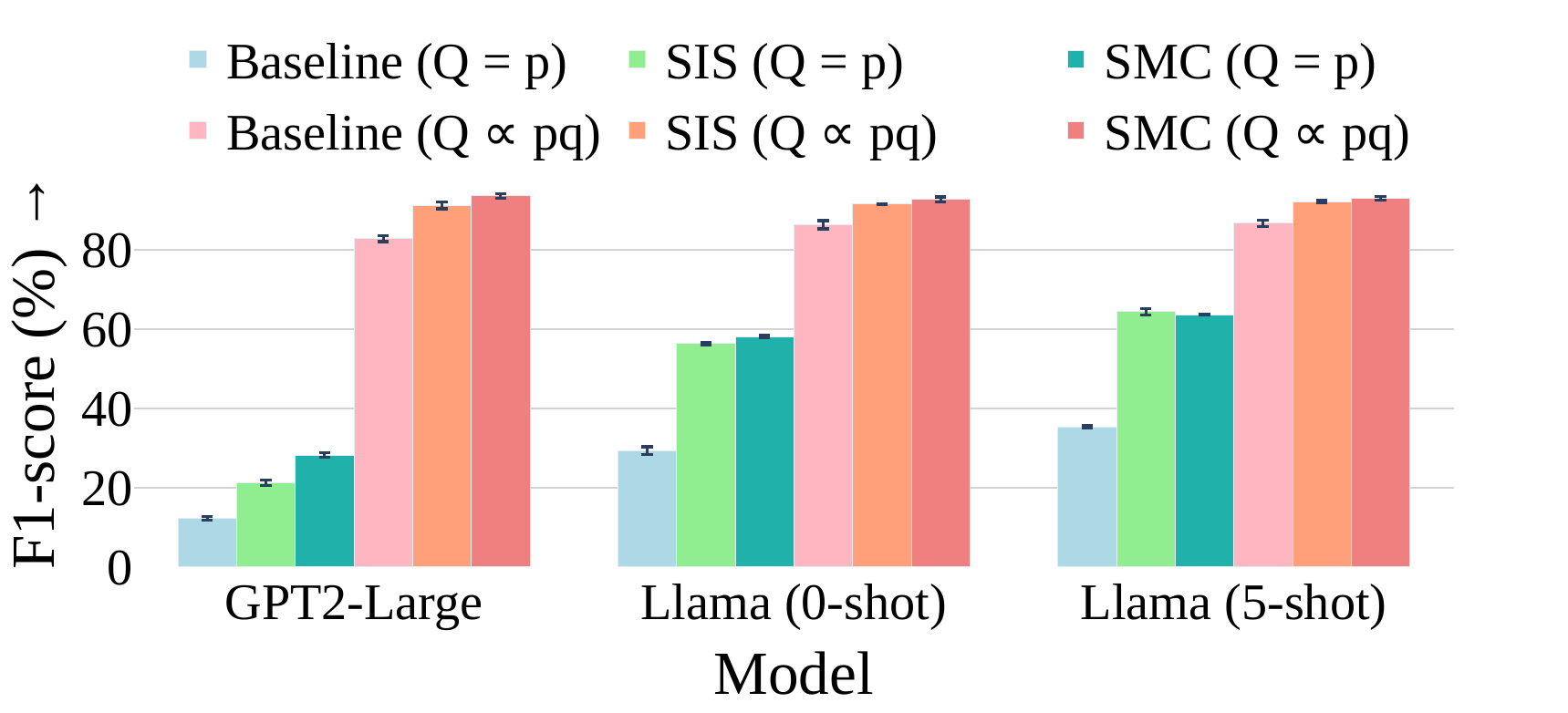}
\caption{F1 score across all models and methods using the two proposal distributions, $\proposal\!=\!\plm$ and $\proposal\!\propto\!\plm\,\ngram$. Note that our baseline is sampling one sentence with $N$ words directly from $\plm$. We observe that \smc achieves the largest F1-score in most cases, while the choice of proposal distribution plays a crucial role in further boosting the syntactic score.
}
\label{fig:f1_score_barplots}
\end{figure}

\subsection{Choice of Proposal Distribution}
In the experiments, we explore two different options for the proposal distribution. The first is the prior distribution itself:
\begin{align}\label{eq:proposal_prior}
\proposal(\sym' \mid \str_{<n}) = 
\plm(\sym' \mid \str_{<n}) 
\end{align}
While simple, the prior may not effectively approximate the posterior $\posterior$, especially when it does not contain much information about the control. 
Therefore, we also consider an alternative proposal that incorporates syntactic information via a bigram model conditioned on part-of-speech tags:
\begin{align}\label{eq:proposal_ngram}
\!\! \proposal(\str' \mid \str_{<n}) \propto 
\plm(\sym_{1}' \mid \str_{<n}) \ngram(\str'\mid \pos_{n}\pos_{n+1}),
\end{align}
where $\sym'_{1}$ is the first token of the word $\str'$,\footnote{This choice was made for efficiency, as evaluating the probability of all possible words is prohibitive.} $\pos_{n}$ is the part-of-speech tag at position $n$ in the syntax tree, and $\ngram$ denotes the probability distribution derived from the bigram model trained on the Penn Treebank dataset.\footnote{More specifically, the bigram model was trained to predict the $n\textsuperscript{the}$ word of a sentence $\str_n$ given the part-of-speech tags $\pos_n$ and $\pos_{n+1}$.  Note that words may be multiple tokens long.}
We chose a bigram model over part-of-speech tags rather than a more sophisticated model, as we found that it adequately captured syntax information without overfitting to its training corpus.
This approach aims to provide a more informed proposal by better approximating the expected future potential in conjunction with the shaping function, which already accounts for the tokens generated so far.

\subsection[Controlled Generation Results]{Controlled Generation Results\protect\footnote{We refer the reader to \Cref{sec:appendix_results} for additional experimental results and supplementary tables.}} 
\label{sec:results}

We repeat each experiment five times for \gptl and two times for \llama models.  The experiments were conducted using $\tau=0.25$ (for \smc), $M\!=\!20$ when $\proposal=\plm$ and $M\!=6\!$ when $\proposal\propto\plm \, \ngram$, as the latter provides a more informed proposal distribution, allowing us to achieve good performance with fewer particles (See \Cref{sec:time}). Note that as $\potential$ we use the Tetratagger trained by \citet{tetra} using the \bert model \citep{devlin2019bertpretrainingdeepbidirectional}, which achieves $95.4$ F1 score. We evaluated all sampling algorithms using language models of varying sizes.

\begin{figure}[t]
\begin{subfigure}{\linewidth}
\centering
\includegraphics[width=\linewidth]{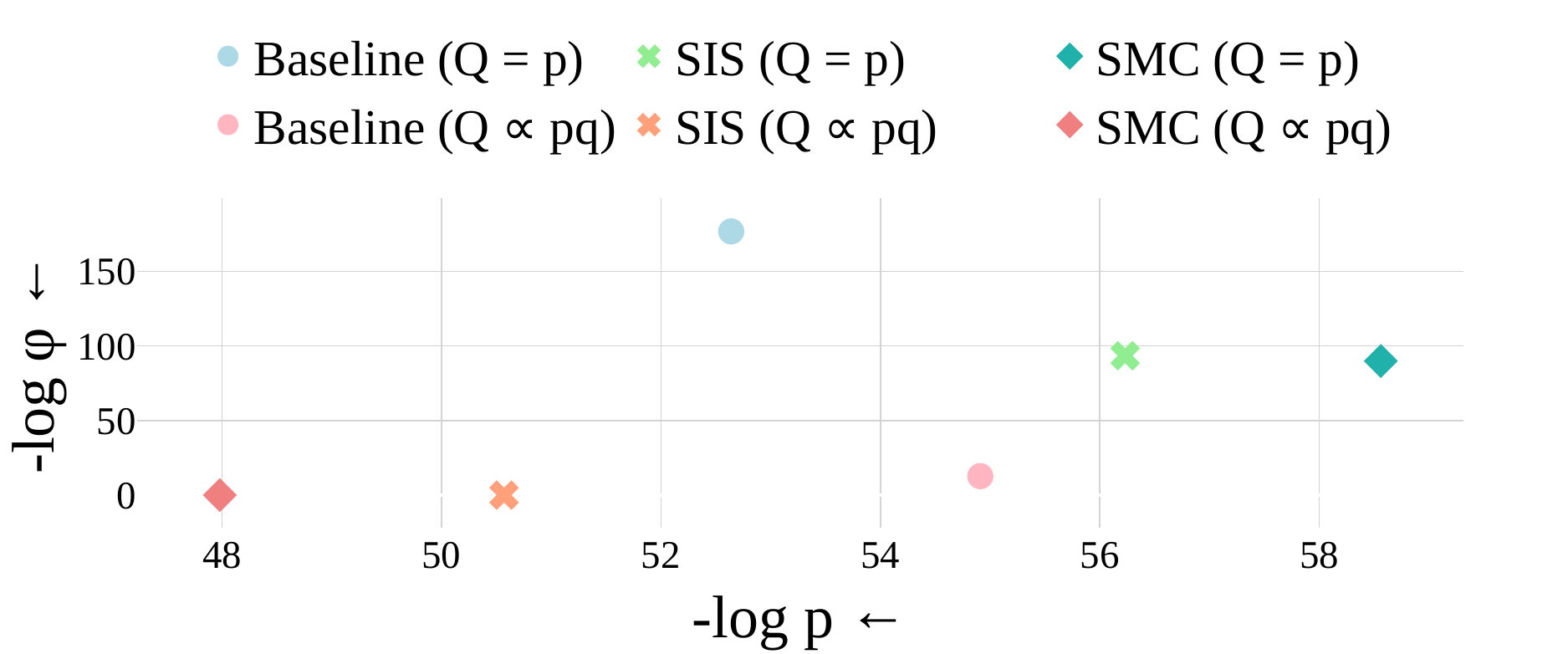}
\caption{\gptl}
\end{subfigure}
\vspace{-0.5em}
\begin{subfigure}{\linewidth}
\centering
\includegraphics[width=\linewidth]{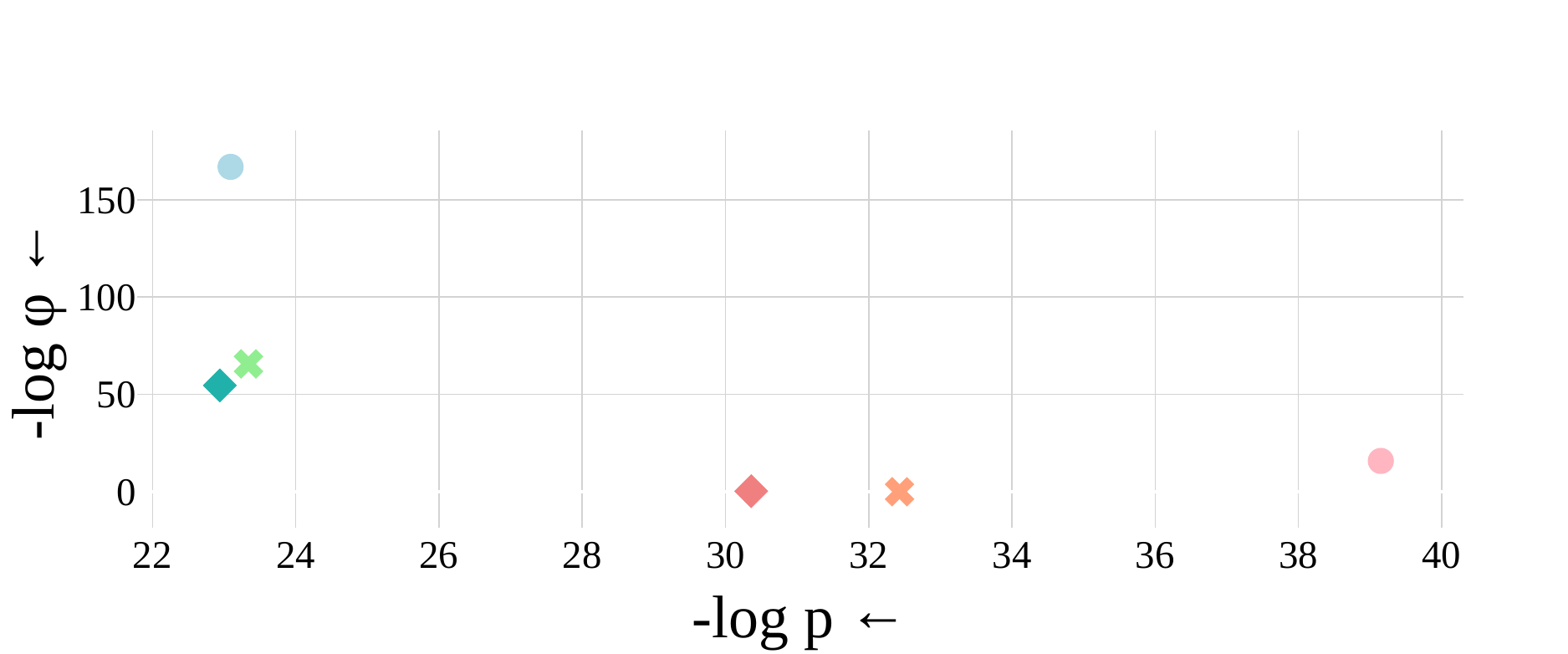}
\caption{\llamai (0-shot)}
\end{subfigure}
\caption{$-\log\potential$ and $-\log\plm$ for different methods, proposal distributions $\proposal=\plm$ and $\proposal=\plm\ngram$ and \gptl and \llamai (0 shots) models. Our approach does not compromise fluency for syntactic consistency. 
}
\label{fig:potential_prior}
\end{figure}

\begin{figure}[t]
\centering
\centering
\includegraphics[width=\columnwidth]{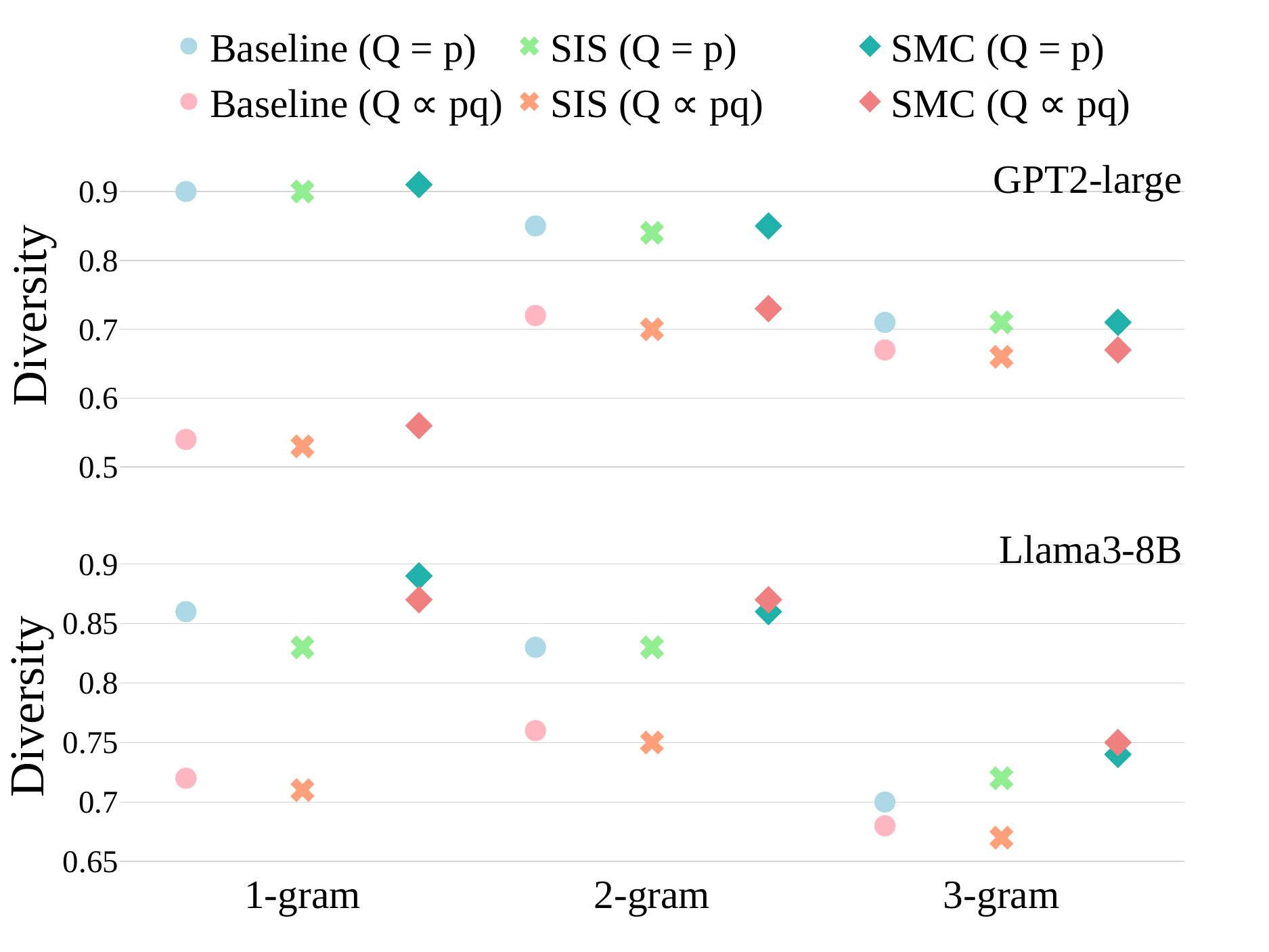}
\caption{Diversity metric across methods and \gptl and \llamai (0 shots) models. $\proposal\propto\plm\,\ngram$ proposal decreases diversity of generated sentences, especially in the case of \gptl model.}
\label{fig:diversity}
\end{figure}

The plot in \Cref{fig:f1_score_barplots} shows that using \Cref{eq:proposal_prior} as the proposal distribution improves F1 scores considerably with both SIS and SMC across all models, more than doubling performance with \smc (from $12.31$ to $28.26$ with \gptl and from $29.41$ to $58.18$ with \llamai (0-shot)). While SMC performs slightly worse than SIS in 5-shot settings, it generally yields higher gains. These findings also highlight the critical role of proposal distribution selection, especially for non-instruction-tuned models. Sampling directly from the prior fails to produce high-likelihood samples, as obtaining a high-quality sample requires $M \propto 1/\Z$, which is impractical when $\Z$ is small. This is further supported by results using $\proposal \propto \plm \, \ngram$, reaching up to $93\%$ F1-score, making smaller models competitive to larger ones. However, this comes at the cost of diversity, particularly with \gptl, where vocabulary restrictions imposed by the bigram model lead to repetitive outputs (See \Cref{fig:diversity}).

We also evaluate syntactic quality and text fluency using $-\log\potential$ and $-\log\plm$, as shown in \Cref{fig:potential_prior}. These metrics capture different aspects of structural fidelity than F1 alone. When using the language model as the proposal distribution, \sis and \smc improve $\log \potential$, achieving a $1/3$ reduction in both \gptl and \llamai, but slightly worsen $\log \plm$, indicating a trade-off. The degradation in $\log\plm$ is more pronounced with \gptl where $-\log\plm$ increases by 5 points, while in \llamai it is marginal. This occurs because our method places greater emphasis on syntactic accuracy; however, the slight degradation shows that it does not compromise the language model’s fluency. Notably, when using $\proposal \propto \plm\, \ngram$, $-\log\potential$ reaches values near zero (i.e., the syntax is nearly a perfect match) with both \smc and \sis, while $-\log\plm$ simultaneously improves. We attribute this to the vocabulary restriction, which may introduce noise when the generation is left uncontrolled, but is effectively corrected when our method is applied.\looseness=-1

\subsection{Decoding Time} 
\label{sec:time}

\begin{figure}[t]
\centering
\includegraphics[width=\columnwidth]{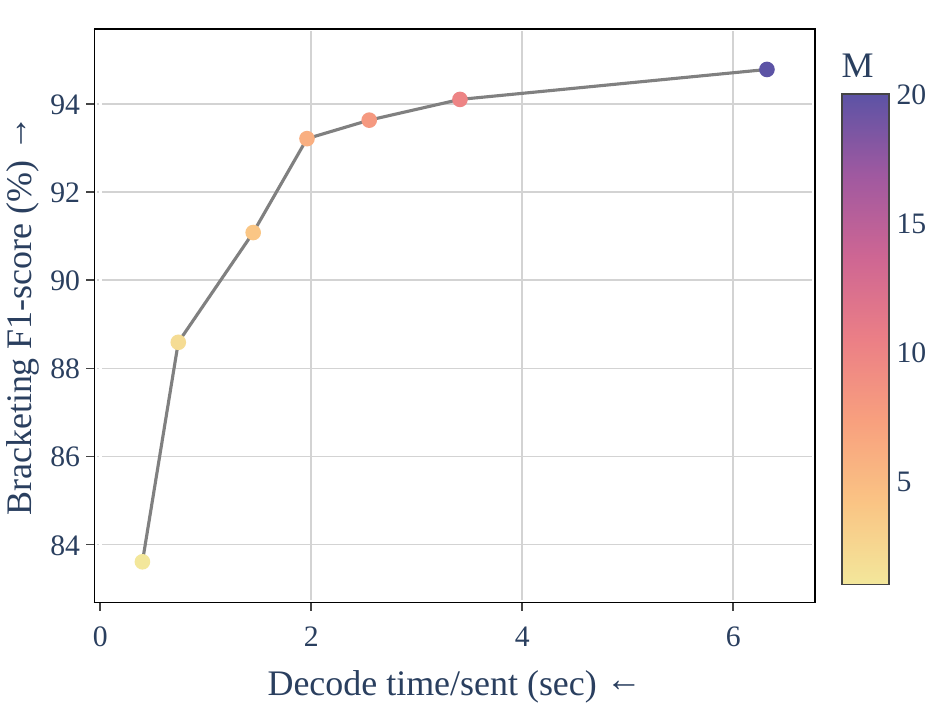}  
\caption{F1-score vs.\@ \smc time/sentence for varying numbers of particles $M$ with $\proposal\propto\plm\,\ngram$ for \gptl. 
}
\label{fig:decode_time}
\end{figure}

\Cref{fig:decode_time} illustrates the relationship between decoding time per sentence and the corresponding bracketing F1 score, for varying values of $M$ using \smc with $\proposal \propto \plm \, \ngram$ on the \gptl model.\footnote{Note that in addition to the time required for sampling from the proposal distribution $\proposal$, all candidate strings must also be evaluated by the tetrataggers $\shape$ and $\potential$.} As $M$ increases, there is a clear improvement in the syntactic quality of the generated sentences, with the F1 score rising from $83\%$ to nearly $95\%$, supporting our hypothesis. However, beyond 6 particles, the improvements become marginal.

\section{Conclusion}

\noindent In this paper, we introduced a sampling algorithm designed to control the syntactic structure of text generated by language models. Our method, based on sequential Monte Carlo, leverages a parsing-as-tagging framework, guiding the generation process by incorporating syntactic taggers. At each step of text generation, the algorithm samples $M$ particles from a proposal distribution and assigns weights to them using a shaping function that assesses the syntactic structure of the partial sequences generated. Our sampling algorithm ensures that both linguistic content and syntactic correctness are taken into account. Our experiments with \gptt and \llama models demonstrate that our approach achieves large improvements in generating text that aligns with target syntactic structures. These results highlight the potential of combining sampling algorithms with tagging frameworks to enhance the syntactic controllability of language models.\looseness=-1 

\section*{Acknowledgments}
The authors would like to thank Ben LeBrun, Manuel de Prada Corral, and Robin Shing Moon Chan for valuable feedback and discussions. Afra Amini is supported by the ETH AI Center doctoral fellowship.

\section*{Limitations}
\paragraph{Pruned-height trees.}
Our method cannot be applied to constituent trees with a pruned height. Since the algorithm relies on full tree structures to assign each token to its corresponding tags, pruning the tree height disrupts this mapping, making the approach unsuitable for such cases.

\paragraph{No fault tolerance.} Our algorithm restricts the length of the generated sentence to the desired number of words defined by the constituent tree. While this ensures that the generated output conforms to specific length constraints, there are instances where a fully coherent sentence is not formed by the time the algorithm finishes. This issue can arise when, in a generation step, the appropriate token does not appear within the $M$ tokens sampled or when the Tetratagger model assigns incorrect tags to words. In such scenarios, if the algorithm could complete the full sentence, even with some syntactic errors, we might see the results would better align with the desired syntax, leading to improved outcomes.\looseness=-1

\paragraph{Language.} In this paper, we focus on English, mainly due to better data availability and improved model performance. However, exploring how this approach generalizes to other languages with diverse syntactic structures is an interesting direction for future work.

\section*{Ethical considerations}
We acknowledge that our approach may introduce unexpected biases or inaccuracies, which could result in outputs that are factually incorrect or contextually inappropriate. This risk arises because our method modifies the probability distribution of a language model to align the generated sentences with specific syntactic structures, which could unintentionally amplify the risk of producing harmful outputs or misinformation. Consequently, users must exercise caution when applying our method, particularly in sensitive or public-facing contexts, to mitigate the potential for unintended negative consequences. 

\bibliography{custom}

\clearpage
\appendix
\onecolumn

\section{Experimental Details}

\subsection{Models}

We experimented with models of different sizes: \gptl ($774$M parameters), \llama ($8$B parameters) for token generation and as the LM backbone of Tetratagger. For \gptf, we used the OpenAI API\footnote{\url{https://openai.com/policies/terms-of-use/}} to generate sentences. All other language models utilized in this research are sourced from HuggingFace \cite{wolf-etal-2020-transformers} and are subject to specific licences that govern their use and distribution in the research domain. Specifically, we used \gptl \footnote{\url{https://huggingface.co/openai-community/gpt2-large}} under MIT License and \llama \footnote{\url{https://huggingface.co/meta-llama/Meta-Llama-3-8B}} and \llamai \footnote{(\url{https://huggingface.co/meta-llama/Meta-Llama-3-8B-Instruct}} under license META LLAMA 3 COMMUNITY LICENSE AGREEMENT.

All experiments for $\gptl$ were conducted on a single \texttt{NVIDIA GeForce RTX 4090 GPU} with $24$ GiB memory, while for $\llama$ on a single \texttt{NVIDIA Tesla V100-SXM2} with $32$ GiB memory.

\subsection{Autoregressive Tetratagger}

We trained our autoregressive Tetratagger models based on \url{https://github.com/nikitakit/tetra-tagging} on the Penn Treebank, as suggested by \citet{tetra}, a standard benchmark for evaluating syntactic parsing algorithms. Our tag vocabulary consists of $231$ labels in total. 

For efficiency reasons, to train a Tetratagger with \llama as the LM backbone, we applied four-bit quantization and fine-tuned the model using LoRA \cite{hu2022lora}. This allows us to reduce the number of trainable parameters to $1.06\%$ of the total model parameters.

\subsection{Evaluation Dataset}
\label{app:dataset}
Our evaluation dataset consists of $301$ syntax trees varying on number of nodes, height, and labels. Dataset statistics are provided in \Cref{tab:data_analysis}. All constituent trees utilized in our experiments were represented in the bracketing representation, which aligns with standard practices in syntactic parsing. 
\begin{table}[h] 
\centering
\begin{tabular}{@{}lcccccc@{}}
\toprule
Number of trees & $301$ \\
Mean height & $7.11$ \\
Max height & $18.00$ \\
Mean leaf nodes &$8.51$ \\
Max leaf nodes & $20.00$ \\
Mean tree size & $24.42$ \\\bottomrule
\end{tabular}
\caption{Statistics for the dataset used for evaluation of our method.}
\label{tab:data_analysis}
\end{table} 
For generating the tag sequences of our syntax trees, we utilized the deterministic algorithm of the pre-trained Tetratagger \citep[\S 3.1;][]{tetra}. In \Cref{fig:tag_seq} we give an example of a tag sequence for a specific tree. For parsing the generated sentences into constituent trees, we used the Berkeley Neural Parser\footnote{\url{https://pypi.org/project/benepar/}} through spaCy.\footnote{\url{https://spacy.io/}} The predicted syntactic trees were then compared to the ground truth syntax trees in our dataset.

\begin{figure}[h] 
\noindent\textbf{Tree}: 
\texttt{(S (NP (EX {\color{PPink}There})) (VP (VBZ {\color{PPink}is}) (ADVP (RB
{\color{PPink}always})) (NP (DT {\color{PPink}a}) (NN {\color{PPink}chance}})))) 

\noindent\textbf{Tag Sequence}: 
\texttt{['l/NP', 'L/S', 'l', 'R/VP', 'l/ADVP', 'R', 'l', 'R/NP', 'r']} 
\caption{Example of a tag sequence given a tree generated by the deterministic algorithm of \citet{tetra}}
\label{fig:tag_seq}
\end{figure}

\subsection{Prompt Formulation}
\label{sec:appendix_prompt}

After applying different prompt formulations for instruction-tuned models, we used the prompts displayed in \Cref{fig:prompts} for $\llamai$ and $\gptf$. For \llamai, we incorporated the special tokens and roles provided by \citet{dubey2024llama3herdmodels} to achieve the best roles. Moreover, we found the appropriate prompt that would force the language model to generate the output sentence directly in the case of \llamai. In the case of \gptf this was not necessary, since we could post-process the generated sentences to remove any explanation or quotation mark.

\begin{figure}
\centering
\begin{subfigure}[t]{0.48\linewidth}
\begin{tcolorbox}[
  enhanced,
  title=Prompt Template,
  separator sign={\tcbline},
  separator sign dash={1pt}{2pt},
  boxrule=0.6pt,
]
\scriptsize
You are a helpful assistant that generates a sentence. From a given constituency parse tree, output only one grammatical English sentence that matches the syntactic structure. Do not include any explanations, preambles, or quotation marks. Respond with the sentence only. \tcbline
An example is first provided: \newline
Parse tree: {\color{PGreen} \{SYNTAX\}} \newline
Sentence: {\color{POrange} \{SENTENCE\}} 
\tcbline
Now generate a sentence for the following tree. \newline
Parse tree: {\color{PGreen} \{SYNTAX\}} 
\tcbline
Sentence: 
\end{tcolorbox}
\caption{\llamai}
\label{fig:prompt_llama3}
\end{subfigure}
\hfill
\begin{subfigure}[t]{0.48\linewidth}
\begin{tcolorbox}[
  enhanced,
  title=Prompt Template,
  separator sign={\tcbline},
  separator sign dash={1pt}{2pt},
  boxrule=0.6pt,
]
\scriptsize
You are a helpful assistant in generating a sentence from the provided 
syntactic structure defined by a constituency parse tree. Please only have the generated sentence, not its parse, in the response. 
\newline
\tcbline
An example is first provided: \newline
Parse tree: {\color{PGreen} \{SYNTAX\}} \newline
Answer: {\color{POrange} \{SENTENCE\}} 
\tcbline
Now generate a sentence for the following parse tree: \newline
Parse tree: {\color{PGreen} \{SYNTAX\}} 
\tcbline
Sentence: 
\end{tcolorbox}
\caption{\gptf}
\label{fig:prompt_gptf}
\end{subfigure}
\caption{The prompt used for generating text with \llamai and \gptf models. Note that the exact format of the prompt has been simplified; in particular, special tokens and roles have been omitted to improve readability.}
\label{fig:prompts}
\end{figure}

\clearpage

\section{Full Experimental Results}
\label{sec:appendix_results}

In \Cref{tab:llmnewresults} we report our full experimental results (the mean and the standard deviation over the runs). Please refer to \Cref{sec:results} for more details about the settings of algorithms \sis and \smc. Note that we observed that $\llama$ tends to generate content related to coding or begins its outputs with questions when prompted with the \bos token. This behavior does not help our method to be applied effectively when the prior equals the language model's distribution.

\begin{table*}[h]
    \centering
    \adjustbox{max width=\textwidth}{%
    \begin{tabular}{@{}lcccccc@{}}
        \toprule
         \textbf{Models} &  \textbf{$\log\potential(\str)$ $\uparrow$} & \textbf{$\log\plm(\str)$ $\uparrow$} & \textbf{F1-score $\uparrow$} &  \multicolumn{3}{c}{\textbf{Diversity}} \\  
         \cmidrule(lr){5-7}
         & & & & \textbf{1-gram $\uparrow$} & \textbf{2-gram $\uparrow$} & \textbf{3-gram $\uparrow$}\\
        \bottomrule
        \multicolumn{7}{c}{$\proposal=\plm$} \\
        \midrule
        \gptl & $-176.80\pmm{9.17}$ & $-52.64\pmm{0.33}$ & $12.31\pmm{0.27}$ & $0.90\pmm{0.05}$ & $0.85\pmm{0.06}$ & $0.71\pmm{0.12}$ \\
        \gptl + \sis & $-93.45\pmm{4.15}$ & $-56.23\pmm{0.78}$ & $21.18\pmm{0.45}$  & $0.90\pmm{0.06}$ & $0.84\pmm{0.06}$ & $0.71\pmm{0.12}$ \\
        \gptl + \smc & $-89.89\pmm{5.58}$ & $-58.56\pmm{0.38}$ & $28.26\pmm{0.39}$ & $0.91\pmm{0.05}$ & $0.85\pmm{0.06}$ & $0.71\pmm{0.12}$ \\
        \noalign{\vspace{7pt}}
        \llama & $-\infty$ & $-62.47\pmm{1.66}$ & $10.05\pmm{0.56}$ & $0.77\pmm{0.23}$ & $0.71\pmm{0.19}$ & $0.61\pmm{0.17}$ \\
        \llama + \sis & $-53.86\pmm{1.78}$ & $-79.79\pmm{1.04}$ & $16.45\pmm{0.24}$ & $0.92\pmm{0.09}$ & $0.85\pmm{0.07}$ & $0.72\pmm{0.12}$ \\
        \llama + \smc & $-64.12\pmm{1.58}$ & $-117.88\pmm{0.33}$ & $16.12\pmm{0.49}$ &  $0.91\pmm{0.09}$& $0.86\pmm{0.05}$ & $0.75\pmm{0.09}$ \\
        \noalign{\vspace{7pt}}
        \llama ($0$-shot) & $-166.86\pmm{4.46}$ & $-23.09\pmm{0.68}$ & $29.41\pmm{0.88}$ & $0.86\pmm{0.09}$ & $0.83\pmm{0.09}$ & $0.70\pmm{0.12}$ \\
        \llama ($0$-shot) + \sis & $-65.66\pmm{0.78}$  & $-23.34\pmm{0.11}$ & $56.42\pmm{0.11}$ & $0.83\pmm{0.10}$ & $0.83\pmm{0.09}$ & $0.72\pmm{0.12}$ \\
        \llama ($0$-shot) + \smc & $-54.58\pmm{1.27}$ & $-22.94\pmm{0.22}$ & $58.18\pmm{0.18}$  & $0.89\pmm{0.07}$ & $0.86\pmm{0.04}$ & $0.74\pmm{0.06}$  \\
        \noalign{\vspace{7pt}}
        \llama ($5$-shot) & $-133.95\pmm{13.20}$  & $-23.12\pmm{0.17}$ & $35.33\pmm{0.11}$ & $0.88\pmm{0.09}$ & $0.84\pmm{0.08}$ & $0.70\pmm{0.12}$ \\
        \llama ($5$-shot) + \sis & $-43.94\pmm{4.62}$   & $-22.71\pmm{0.36}$ & $64.44\pmm{0.67}$ & $0.86\pmm{0.09}$& $0.84\pmm{0.09}$& $0.72\pmm{0.12}$ \\
        \llama ($5$-shot) + \smc & $-40.89\pmm{5.51}$  & $-24.02\pmm{0.09}$  & $63.66\pmm{0.01}$ & $0.92\pmm{0.06}$ & $0.87\pmm{0.04}$ & $0.74\pmm{0.06}$ \\
        \midrule
        \multicolumn{7}{c}{$\proposal\propto\plm\,\ngram$} \\
        \midrule
        \gptl & $-12.71\pmm{1.17}$ & $-54.91\pmm{0.53}$ & $82.89\pmm{0.61}$ &  $0.54\pmm{0.09}$ & $0.72\pmm{0.10}$ & $0.67\pmm{0.12}$\\
        \gptl + \sis & $-0.0001\pmm{0.0}$ & $-50.57\pmm{0.16}$ & $91.22\pmm{0.66}$  & $0.53\pmm{0.09}$ & $0.70\pmm{0.11}$ & $0.66\pmm{0.13}$ \\
        \gptl + \smc & $-0.0009\pmm{0.0001}$ & $-47.98\pmm{0.34}$ & $93.69\pmm{0.33}$ & $0.56\pmm{0.09}$ & $0.73\pmm{0.11}$  & $0.67\pmm{0.13}$ \\
        \noalign{\vspace{7pt}}
        \llama & $-14.93\pmm{0.49}$ & $-89.11\pmm{0.24}$ & $76.49\pmm{0.88}$ &  $0.75\pmm{0.09}$& $0.81\pmm{0.09}$& $0.71\pmm{0.12}$ \\
        \llama + \sis & $-0.28\pmm{0.20}$  & $-75.58\pmm{0.24}$  & $86.50\pmm{0.11}$ & $0.75\pmm{0.09}$ & $0.81\pmm{0.08}$ & $0.71\pmm{0.12}$ \\
        \llama + \smc & $-6.68\pmm{0.07}$ & $-70.80\pmm{0.87}$  & $90.18\pmm{0.18}$ & $0.88\pmm{0.07}$ & $0.86\pmm{0.04}$ & $0.73\pmm{0.08}$ \\
        \noalign{\vspace{7pt}}
        \llama ($0$-shot) & $-15.83\pmm{0.41}$ & $-39.16\pmm{0.43}$  & $86.38\pmm{1.07}$ & $0.72\pmm{0.10}$ & $0.76\pmm{0.11}$& $0.68\pmm{0.14}$ \\
        \llama ($0$-shot) + \sis & $-0.002\pmm{0.001}$&  $-32.44\pmm{0.15}$ & $91.58\pmm{0.01}$ & $0.71\pmm{0.11}$ & $0.75\pmm{0.13}$ & $0.67\pmm{0.15}$ \\
        \llama ($0$-shot) + \smc & $-0.28\pmm{0.24}$ & $-30.36\pmm{0.68}$  & $92.78\pmm{0.35}$ & $0.87\pmm{0.06}$  & $0.87\pmm{0.04}$ & $0.75\pmm{0.07}$ \\
        \noalign{\vspace{7pt}}
        \llama ($5$-shot) & $-8.02\pmm{4.62}$ &  $-37.17\pmm{0.21}$& $86.77\pmm{0.62}$  &  $0.75\pmm{0.10}$&$0.79\pmm{0.10}$ & $0.70\pmm{0.13}$ \\
        \llama ($5$-shot) + \sis & $-0.002\pmm{0.0}$ & $-31.87\pmm{0.42}$ & $92.24\pmm{0.16}$ & $0.74\pmm{0.10}$  & $0.79\pmm{0.11}$ & $0.69\pmm{0.14}$ \\
        \llama ($5$-shot) + \smc & $-0.001\pmm{0.002}$  & $-29.96\pmm{0.24}$ & $93.05\pmm{0.28}$ & $0.88\pmm{0.06}$  & $0.87\pmm{0.04}$ & $0.75\pmm{0.07}$ \\
        \bottomrule
    \end{tabular}}
    \caption{Full table of results.  Please refer to \cref{sec:results} for plots and discussion.
    }
    \label{tab:llmnewresults} 
\end{table*} 

\clearpage
\section{Notation Glossary}
\begin{table*}[h]
\centering
\begin{tabular}{@{}ll@{}}
\toprule
symbol & meaning \\ \midrule
$N$ & string length (number of tokens) \\
$\eos$ & end-of-string marker \\ 
$\sym,\sym' \in \alphabet$ & symbols in the alphabet $\alphabet$ \\
$\str \in \alphabet^*$ & string from the set of string $\alphabet^*$ \\
$\str_n \in \alphabet$ & $n\textsuperscript{th}$ symbol in string $\str$ \\
$\alphabet$ & alphabet; set of tokens \\
$\alphabet^*$ & set of all strings made from symbols in $\alphabet$ \\
\midrule
$L$ & string length (number of words) \\
$\word$ & word, defined as a sequence of tokens and identified as a single word by spaCy \\
\midrule
$\bt$ & target syntax tree \\
$\posterior(\str\mid\bt)$ & posterior over strings \\
$\posterior(\str)$ & posterior over strings; abbreviation for $\posterior(\str\mid\bt)$ when $\bt$ is clear from context \\
$\plm(\str)$ & language model prior of strings \\
$\potential(\bt \mid \str)$ & likelihood; the probability that $\str$ has the syntax tree $\bt$ \\
$\potential(\str)$ & likelihood; abbreviation for $\potential(\bt \mid \str)$ when $\bt$ is clear from context \\
$\Z(\bt)$ & posterior normalization constant \\
$\Z$ & posterior normalization constant; abbreviation for $\Z(\bt)$ when $\bt$ is clear from context \\
\midrule
$M$ & sample size \\
$\proposal(\str)$ & proposal distribution \\
$\shape(\str)$ & shaping function; autoregressive tetratagger \\
$\pest(\str)$ & posterior approximation \\
$\zest$ & estimated normalization constant \\
$\uweight$ & importance weight \\
\midrule
$\ngram(\str \mid \pos_n, \pos_{n+1})$ & bigram proposal \\ 
$\pos_n$ & part-of-speech tag for the $n\textsuperscript{th}$ word of $\bt$ \\
\midrule
$\tags$ & set of tags (tetratagger) \\
$\lm(\word_\ell \mid \str) \in \mathbb{R}^d$ & transformer language model hidden state with dimensionality $d$ \\
$\projOdd, \projEven \in \mathbb{R}^{\tags \times d}$ & tetratagger parameters \\
\bottomrule
\end{tabular}
\caption{Notation glossary}
\label{tab:notation}
\end{table*}

\end{document}